\documentclass{bmvc2k}

\usepackage{multirow}
\usepackage{textcomp}
\usepackage{amssymb}

\title{FacialGAN: Style Transfer and Attribute Manipulation on Synthetic Faces}

\addauthor{Ricard Durall}{ricard.durall.lopez@itwm.fraunhofer.de}{1,2}
\addauthor{Jireh Jam}{jireh.jam@stu.mmu.ac.uk}{3}
\addauthor{Dominik Strassel}{dominik.strassel@itwm.fraunhofer.de}{1}
\addauthor{Moi Hoon Yap}{m.yap@mmu.ac.uk}{3}
\addauthor{Janis Keuper}{janis.keuper@hs-offenburg.de}{1,4}

\addinstitution{
 Fraunhofer ITWM,\\
 Kaiserslautern, Germany
}
\addinstitution{
 IWR, University of Heidelberg,\\
 Heidelberg, Germany
}
\addinstitution{
 Manchester Metropolitan University,\\
 Manchester, United Kingdom
}
\addinstitution{
 IMLA, Offenburg University,\\
 Offenburg, Germany
}

\runninghead{Durall, Jam, Straßel, Yap, Keuper}{FacialGAN}

\begin{document}

\maketitle

\begin{abstract}
Facial image manipulation is a generation task where the output face is shifted towards an intended target direction in terms of facial attribute and styles. 
Recent works have achieved great success in various editing techniques such as style transfer and attribute translation.
However, current approaches are either focusing on pure style transfer, or on the translation of predefined sets of attributes with restricted interactivity. 
To address this issue, we propose FacialGAN, a novel framework enabling simultaneous rich style transfers and interactive facial attributes manipulation.
While preserving the identity of a source image, we transfer the diverse styles of a target image to the source image. We then incorporate the geometry information of a segmentation mask to provide a fine-grained manipulation of facial attributes.  
Finally, a multi-objective learning strategy is introduced to optimize the loss of each specific tasks.
Experiments on the CelebA-HQ dataset, with CelebAMask-HQ as semantic mask labels, show our model’s capacity in producing visually compelling results in style transfer, attribute manipulation, diversity and face verification. 
For reproducibility, we provide an interactive open-source tool to perform facial manipulations, and the Pytorch implementation of the model.

\end{abstract}

\begin{figure}[ht]
    \includegraphics[width=\linewidth]{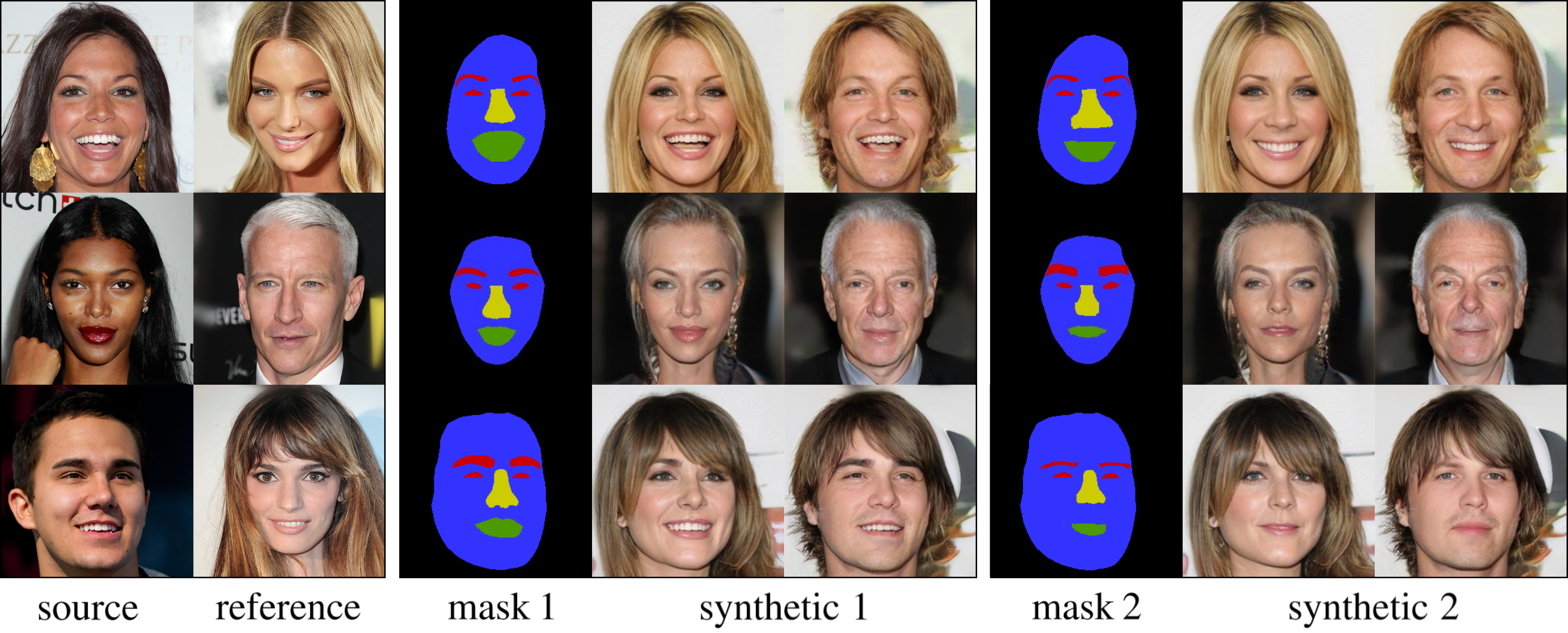}
   \caption{Given a source and a reference image, users can change the mask of the source so that the synthetic result retains the identity from the source, with the style from the reference and follows the customized mask (e.g., modified eyes, eyebrows, nose or mouth).}
\label{fig:teaser}
\end{figure}

\section{Introduction}
Facial image manipulation is a challenging task that involves generating images whilst preserving the subtle texture of relevant features of the faces. 
When editing, different levels of structural changes are imposed on key characteristics so that the system steers towards the target direction, with the attempt to synthesize realistic facial images.
Generative neural networks have been very successful in this task, due to their ability to extract high-level features, and they have been utilized in different editing tasks.
Common editing techniques like image-to-image translation include style transfer \cite{isola2017image,choi2020stargan} and attribute manipulation \cite{choi2018stargan,lee2020maskgan}.

Style transfer approaches define two or more visual domains, where the goal is to translate from one  domain to another.
Usually, these domains represent distinguishable properties like gender, hairstyle, and skin colour among others.  
Ideally, while training these models, one needs to predefine the boundaries of the target domains/styles, as they can be arbitrarily large. 
To address this problem, \cite{choi2020stargan} introduces an approach, where the styles are controlled either by 
domain specific encoders or by semantic labels.
This structure allows to have an image-to-image translation model that can handle a wide diversity of styles.
Despite the remarkable results, this model has two important limiting factors.
First, the scalability over multiple domains has a direct impact on the size of the architecture.
Additionally, it might require a pre-classification of the data according to the target domain, which is not always a simple task, since some domains cannot be binarised. 
The second limitation comes from its semantic manipulation nature, i.e., it translates whole images, not allowing local nor pixel-wise manipulations.
As a consequence, it cannot have a fine-grained control of the faces, generating less diverse attributes than a geometry manipulation approach.

With facial attribute geometry manipulations, we can achieve rich generation of attributes thanks to its semantic reasoning. 
It consists of an image generation process, closely bounded to the aforementioned technique, with supervision from a well-defined feature (target attribute), which is modified with consistency preserving realism with the rest of the face. 
We can categorize attribute manipulation techniques into semantic-level manipulation \cite{liu2017unsupervised,lample2017fader,choi2018stargan} and geometry-level manipulation \cite{xiao2018elegant,yin2019instance,park2019semantic}. 
The former is precise and easy to train, however, it does not allow users to interactively manipulate the face images. 
Recent works on geometry manipulation \cite{gu2019mask,lee2020maskgan,wei2020maggan} have achieved notable results using semantic masks as intermediate representations of facial features. 
While it is true that these methods grant the user with more freedom to manipulate the attributes at will, they are not generating high diversity outputs.
\cite{lee2020maskgan} overcomes these drawbacks by mimicking user's manipulation via a mask manifold.
Nonetheless, this method is limited in terms of style transfer, since it lacks the ability to conduct morphological changes when applying style. 
A possible solution to mitigate this issue is describing the style via the available attributes, but it is quite restricted.

To address these limitations, we propose a network that learns both styles and semantic attribute translation, thereby ensuring that it can deal with all the aforementioned tasks at once. 
Similar to \cite{choi2020stargan}, we design a network that extracts and applies diverse styles to generate realistic facial images. Additionally, inspired by  \cite{lee2020maskgan}, we incorporate geometry information via segmentation masks to achieve a fine-grained manipulation of facial attributes, leading to a rich diversity of outputs.
We believe that having a functional end-user application is important to both industry and academia, hence, we provide an open-source tool similar to \cite{faceapp}, where the user can play around and experiment with our pretrained model\footnote {Pytorch implementation is available at the \href{https://github.com/cc-hpc-itwm/FacialGAN}{https://github.com/cc-hpc-itwm/FacialGAN}}.

\vspace{-.5mm}
\section{Related Work}


Generative Adversarial Networks (GANs) \cite{goodfellow2014generative} have shown impressive results in various computer vision tasks like image generation \cite{brock2018large,karras2020analyzing}, 
image-to-image translation \cite{isola2017image,durall2021local},
inpainting \cite{yu2019free,jam2021r}
and image segmentation \cite{Kalayeh_2017_CVPR,benini2019fasseg}.
Among them, facial manipulation tasks have continuously gained attention in recent years due to the high demand of facial editing applications.
Facial manipulation can be seen as a multi-domain image-to-image translation problem, where the model works with unique domain features from the face. 
From a style transfer point of view \cite{isola2017image,zhu2017unpaired,liu2017unsupervised,chen2017photographic,wang2018high,lee2018diverse,mao2019mode,choi2020stargan}, each domain consists of styles, e.g., hairstyle, makeup and skin colour.
From an attribute translation view \cite{xiao2017dna,perarnau2016invertible,choi2018stargan,he2019attgan,wu2019relgan,xiao2018elegant,park2019semantic,chen2019semantic,wei2020maggan,lee2020maskgan}, each domain consists of attributes, e.g., smiling, beard and big nose.

One of the pioneer works on style translation \cite{isola2017image} suggested learning to map from a source to a target domain using paired images in a supervised manner.
Shortly after, new techniques such as cycle consistency \cite{zhu2017unpaired} or shared latent space \cite{liu2017unsupervised} were introduced to remove the need for pairs, reducing in this way the dataset complexity.
Concurrently, more advanced topologies were developed, e.g., cascaded refinement \cite{chen2017photographic} or multi-scale \cite{wang2018high}.
Follow-up works proposed to use disentangled representations based on a domain-invariant content space and a domain-specific attribute space \cite{lee2018diverse}, on a maximization of the ratio of the distance between generated images and their latent codes \cite{mao2019mode}, or on a framework that tackled both diversity of generated images and scalability over multiple domains \cite{choi2020stargan}.

Attribute translation tasks have also undergone major changes -- including semantic-level and geometry-level manipulations.
Under a semantic-level umbrella,
\cite{xiao2017dna} tried to approach the attribute translation task, by generating swapping attribute-related blocks in the latent space between two images.
\cite{perarnau2016invertible} combined a conditional GAN with an encoder, which allowed manipulating multiple attributes at once. 
\cite{choi2018stargan} introduced an important breakthrough by employing a single generator to perform multi-domain image translation. 
\cite{he2019attgan} also achieved remarkable results with an encoder-decoder architecture, where the attribute information has been treated as a part of the latent representation.
For a finer control, \cite{wu2019relgan} primarily took advantage of relative attributes, which described the desired change on selected attributes.
As for geometry-level attribute approaches, \cite{xiao2018elegant} introduced a new model that used two images of opposite attributes as inputs, to transfer exactly the same type of attributes from one image to another by exchanging certain part of their encodings.
A different approach followed \cite{park2019semantic}, where they employed
the input layout for modulating the activations in normalization layers through a spatially-adaptive, learned transformation.
\cite{chen2019semantic} suggested to decompose facial attributes into multiple semantic components, each corresponding to a specific face region.
Similarly, \cite{wei2020maggan} incorporated the influence region of each attribute into the generator, and they combined it with a multi-level patch-wise discriminator structure.
Finally, \cite{lee2020maskgan} enabled diverse and interactive face manipulation via semantic masks that served as intermediate representations for flexible face manipulations with fidelity preservation.

\section{Contributions}
In this work, we focus on the challenging task of facial image editing.
Given an input face image, a target style face image, and a guidance segmentation label mask, our novel framework is able to synthesize an output image that (1) shares a similar style with the target style image, while preserving the input face identity, and (2) follows accurately the semantic mask.
To the best of our knowledge, FacialGAN is a pioneer of incorporating both techniques under the same umbrella, leading to a flexible and fine-grained editing control.
As a result, our facial system achieves state-of-the-art scores in reference-guided synthesis, improving seminal works such as \cite{choi2020stargan,lee2020maskgan}.
We propose a multi-objective training that is able to balance the different components of the architecture.
In particular, we introduce a new local segmentation loss to encourage the network to follow the geometry specified in the guidance face mask.
Unlike \cite{lee2020maskgan} where a complex
training strategy generates a supervision signal, our segmentation loss back-propagates informative gradients thanks to its locality characteristics.
In other words, it exploits the region of interest, i.e., the target pixel-wise attributes.
Overall, our contributions are summarized as follows:

\begin{itemize}
\item  We propose a novel model enabling simultaneous rich style transfers and interactive facial attributes manipulation, while maintaining the identity.

\item We introduce an intuitive local segmentation loss that guarantees the pixel-wise attribute control, simplifying the complex global pipelines of \cite{lee2020maskgan}.

\item We assess both qualitatively and quantitatively results on CelebA-HQ dataset.
We report state-of-the-art scores on reference-guided generation, surpassing \cite{choi2020stargan,lee2020maskgan}.

\item We provide an open-source framework which enables the user through an interactive GUI to manipulate the structure of facial features along with transferring styles.
\end{itemize}

\section{Our Approach}

\subsection{Problem Formulation}
Given a source image $\mathrm{\mathbf{x}}\in \mathbb{R}^{H \times W \times 3}$, its segmentation label mask $\mathrm{\mathbf{m}}\in \mathbb{R}^{H \times W \times C}$, and an arbitrary reference image $\mathrm{\mathbf{y}}\in \mathbb{R}^{H \times W \times 3}$, our goal is to train a model that can transfer the style from $\mathrm{\mathbf{y}}$ to $\mathrm{\mathbf{x}}$, control the gender domain, and being consistent with the geometry constraint of $\mathrm{\mathbf{m}}$.
Note that $C$ is the category number of the semantic label.

To obtain images with diverse styles and flexible attribute manipulation, we train our approach on a twofold task: (1) to generate domain-specific style vectors from arbitrary images and random noise, and (2) to synthesize realistic faces following the geometry dictated by pseudo-random masks.
By enforcing such a behaviour, the model will learn to reflect the style vectors and the changes on the attributes of the masks, producing images with diversity and scalability over multiple domains.

\begin{figure}
\centering
    \includegraphics[width=\linewidth]{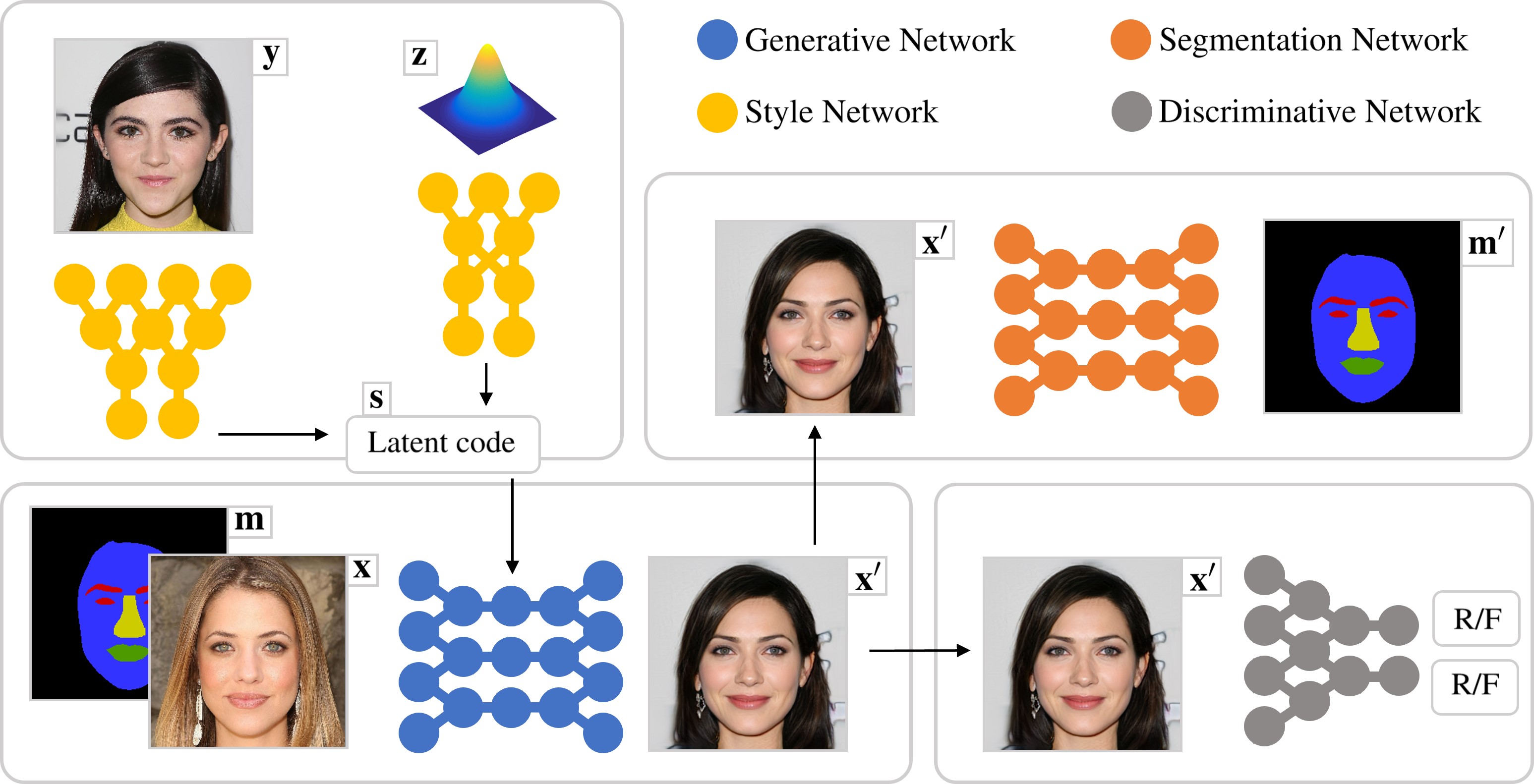}
   \caption{Pipeline of FacialGAN framework. 
   Given an input image $\mathrm{\mathbf{x}}$, the generator synthesizes a new face $\mathrm{\mathbf{x'}}$, conditioned on a latent code $\mathrm{\mathbf{s}}$ and a semantic mask $\mathrm{\mathbf{m}}$. 
   The style network extracts the $\mathrm{\mathbf{s}}$ containing the style.
   It can be defined either from a reference image $\mathrm{\mathbf{y}}$, or from random noise $\mathrm{\mathbf{z}}$. 
   To ensure that the generated face is consistent with the mask, we employ a segmentation network. 
   Finally, the discriminator classifies the output: real (R) or fake (F).}
\label{fig:pipeline}
\end{figure}

\subsection{Model Architecture}
The topology of our proposal is depicted in \autoref{fig:pipeline}.
It is composed of a generative network, a style network, a segmentation network and a discriminative network. 
By combining each of these blocks sequentially, the ensemble model successfully transfers styles and attributes.
\noindent \textbf{Generative Network.}
The task of the generator $G$ is to translate an input image $\mathrm{\mathbf{x}}$ into an output image $\mathrm{\mathbf{x'}}$, following the label mask $\mathrm{\mathbf{m}}$, while reflecting the style code $\mathrm{\mathbf{s}}$.
Inspired by \cite{durall2021local}, we use an encoder-decoder topology and randomly mask one of the attributes of $\mathrm{\mathbf{x}}$ so that the network learns to inpaint coherent attributes.
Then, we concatenate the masked image with the mask $\mathrm{\mathbf{m}}$, and feed it into the encoder.
For the style transfer, we inject $\mathrm{\mathbf{s}}$ into the decoder using adaptive instance normalization \cite{huang2017arbitrary}.
\noindent \textbf{Style Network.}
The aim of this network is to generate valid style codes $\mathrm{\mathbf{s}}$.
To do that, the network is split into two subnetworks: a mapping network $F$ and an encoder network $E$. 
While $F$ generates a style code from random noise $\mathrm{\mathbf{z}}$, $E$ extracts the style from input images.
We adopt the architecture of \cite{choi2020stargan} as a backbone and simplify it for a binary domain to control the gender information. The remaining attributes are manipulated through the mask.
\noindent \textbf{Segmentation Network.}
To guarantee a diverse and interactive face manipulation, we need to ensure that $G$ follows the geometry of the mask.
Therefore, the segmentation network $S$ generates a control signal that penalizes the generator the moment that the output $\mathrm{\mathbf{x'}}$ and $\mathrm{\mathbf{m}}$ are not aligned.
To achieve that, we feed $\mathrm{\mathbf{x'}}$ into $S$ and compare the generated output mask $\mathrm{\mathbf{m'}}$ with the label mask $\mathrm{\mathbf{m}}$ which serve as ground-truth.
\noindent \textbf{Discriminative Network.}
The last component is a convolutional discriminator $D$. 
However, it behaves slightly different from the vanilla implementation \cite{goodfellow2014generative}, as it takes samples of both real and generated faces and tries to correctly classify them into real and fake based on the gender domain. 
This discrimination procedure is called ``multi-task classification'' and it has been successfully employed in previous works \cite{liu2019few,choi2020stargan}.

\subsection{Multi-Objective Learning}
Learning to synthesize realistic and diverse images while transferring styles and manipulating attributes is a complex task.
It requires different regularizer terms that focus on specific tasks.
In this work, we mainly use five independent losses in training to achieve our goal. 
\begin{equation}
    \mathcal{L}_{\mathrm{adv}} = \mathbb{E}_{\mathrm{\mathbf{x}}} [\log D(\mathrm{\mathbf{x}})]  + \mathbb{E}_{\mathrm{\mathbf{x}},\mathrm{\mathbf{m}},\mathrm{\mathbf{z}}} [\log(1-D(G(\mathrm{\mathbf{x}},\mathrm{\mathbf{m}},F(\mathrm{\mathbf{z}}))))]
\end{equation}
The adversarial loss $\mathcal{L}_{\mathrm{adv}}$ \cite{goodfellow2014generative} is the core element in any GAN-based model.
Essentially, it makes the generated images more realistic and assess the control over gender's domain. 
\begin{equation}
    \mathcal{L}_{\mathrm{sty}} = \mathbb{E}_{\mathrm{\mathbf{x}},\mathrm{\mathbf{m}},\mathrm{\mathbf{z}}} [||\mathrm{\mathbf{s}} - E(G(\mathrm{\mathbf{x}},\mathrm{\mathbf{m}},\mathrm{\mathbf{s}}))||_1]
\end{equation}
The style loss $\mathcal{L}_{\mathrm{sty}}$ \cite{choi2020stargan} is vital to achieve reliable style transfers. 
It is responsible to enforce the generator to utilize the style codes $\mathrm{\mathbf{s}}$, extracted from $F(\mathrm{\mathbf{z}})$, by  minimizing the distance between them and the style codes, extracted from $E$ when feeding with generated images. 
\begin{equation}
    \mathcal{L}_{\mathrm{ds}} = - \mathbb{E}_{\mathrm{\mathbf{x}},\mathrm{\mathbf{m}},\mathrm{\mathbf{z_1}},\mathrm{\mathbf{z_2}}} [||G(\mathrm{\mathbf{x}},\mathrm{\mathbf{m}},F(\mathrm{\mathbf{z_1}})) - G(\mathrm{\mathbf{x}},\mathrm{\mathbf{m}},F(\mathrm{\mathbf{z_2}}))||_1]
\end{equation}
By maximizing the distance between two generated images with respect to their corresponding latent codes $\mathrm{\mathbf{z_1}}$ and $\mathrm{\mathbf{z_2}}$, the diverse sensitivity loss $\mathcal{L}_{\mathrm{ds}}$ \cite{mao2019mode} forces the generator to explore more minor modes and therefore, to produce more diversity. 
\begin{equation}
    \begin{split}
    \mathcal{L}_{\mathrm{cyc}} = & \mathbb{E}_{\mathrm{\mathbf{x}},\mathrm{\mathbf{m}},\mathrm{\mathbf{z}}} [||\mathrm{\mathbf{x}} - G(\mathrm{\mathbf{x'}},\mathrm{\mathbf{m}},E(\mathrm{\mathbf{x}}))||_1] \\
    & \mathrm{with} \;\; \mathrm{\mathbf{x'}} = G(\mathrm{\mathbf{x}},\mathrm{\mathbf{m}},F(\mathrm{\mathbf{z}}))
    \end{split}
\end{equation}
The cyclic consistency loss $\mathcal{L}_{\mathrm{cyc}}$ \cite{zhu2017unpaired} guarantees the preservation of the domain invariant characteristics (e.g., pose), while changing its styles faithfully.
\begin{equation}
    \begin{split}
    \mathcal{L}_{\mathrm{seg}} = -\sum_{h,w} m^{h, w,c} \log S(\mathrm{\mathbf{x}}^{h,w,c}) + (1-m^{h,w,c}) \log (1-S(\mathrm{\mathbf{x}}^{h,w,c}))
    \end{split}
\end{equation}
Finally, the local segmentation loss $\mathcal{L}_{\mathrm{seg}}$ is based on binary cross-entropy with the singularity that it works locally.
Depending on the manipulated attribute $c$, $\mathcal{L}_{\mathrm{seg}}$ will evaluate a certain image region ($h$,$w$). 
The goal of this loss is to ensure that the mask rules the attribute geometry of output images. The overall objective can be formulated as
\begin{equation}
    \min_{G,F,E,S} \max_{D} \mathcal{L}_{\mathrm{final}} = 
    \lambda_{\mathrm{adv}} \mathcal{L}_{\mathrm{adv}}  \,+\, \lambda_{\mathrm{sty}} \mathcal{L}_{\mathrm{sty}}  \,+\, 
    \lambda_{\mathrm{ds}} \mathcal{L}_{\mathrm{ds}}  \,+\, 
    \lambda_{\mathrm{cyc}} \mathcal{L}_{\mathrm{cy}}  \,+\, 
    \lambda_{\mathrm{seg}} \mathcal{L}_{\mathrm{seg}},
\end{equation}
where $\lambda_{\mathrm{adv}}$, $\lambda_{\mathrm{sty}}$, $\lambda_{\mathrm{ds}}$, $\lambda_{\mathrm{cyc}}$ and $\lambda_{\mathrm{seg}}$ are the hyperparameters for each term.

\section{Experiments}

\subsection{Experimental Setup}

\noindent \textbf{Baselines Models.}
We choose state-of-the-art DRIT \cite{lee2018diverse}, MSGAN \cite{mao2019mode}, SPADE \cite{park2019semantic}, StarGANv2  \cite{choi2020stargan} and MaskGAN \cite{lee2020maskgan} as our baselines for comparison.
DRIT, MSGAN and StarGANv2 perform latent-guided and reference-guided style transfer.
Whereas SPADE and MaskGAN perform geometry-level facial attribute manipulation.
\noindent \textbf{Datasets.}
We use CelebA-HQ \cite{karras2017progressive} and CelebAMask-HQ \cite{lee2020maskgan} datasets. 
While CelebA-HQ contains 30,000 high-quality facial images picked from the CelebA \cite{liu2015deep} dataset, CelebAMask-HQ contains the corresponding semantic segmentation labels separated on 19 classes.
For our experiments we resize all images to the size of 256$\times$256, and we create and employ four customized classes -- eyes, nose, mouth and skin.
\noindent \textbf{Training Details.}
First, we train our segmentation model based on \cite{ronneberger2015u}, for 50 epochs using a batch size of 32, with the default Adam \cite{kingma2014adam} optimizer with learning rate set to $10^{-2}$. 
Then, we train our generative model for 200,000 iterations using a batch size of 8, but in this training we have four Adam optimizers, where we set $\beta_{1} = 0$ and $\beta_{2} = 0.99$ with learning rates $10^{-4}$ for $G$, $D$ and $E$, and $10^{-6}$ for $F$. 
The losses are all equally weighted except for the segmentation with $\lambda_{\mathrm{seg}}$ = 2, and the style diversification where $\lambda_{\mathrm{ds}}$ is linearly decayed to zero over training. 

\begin{table}[t]
    \begin{minipage}[b]{0.5\linewidth}
		\centering
        \begin{tabular}[b]{|l|c|c|}
        \hline
        Method & FID $\downarrow$ & LPIPS $\uparrow$ \\
        \hline\hline
        DRIT \cite{lee2018diverse} & 52.1 &  0.178 \\
        MSGAN \cite{mao2019mode} & 33.1 &  0.389  \\
        StarGANv2 \cite{choi2020stargan} & \textbf{13.7} & \textbf{0.452} \\
        Ours & 15.8 & 0.426 \\
        \hline
        \end{tabular}
	\end{minipage}\hfill
    \begin{minipage}[b]{0.5\linewidth}
        \centering
        \begin{tabular}[b]{|l|c|c|}
        \hline
        Method & FID $\downarrow$ & LPIPS $\uparrow$ \\
        \hline\hline
        DRIT \cite{lee2018diverse} & 53.3 & 0.311 \\
        SPADE \cite{park2019semantic} & 46.2 & - \\
        MSGAN \cite{mao2019mode} & 39.6 & 0.312 \\
        MaskGAN \cite{lee2020maskgan} & 37.1 & - \\
        StarGANv2 \cite{choi2020stargan} & 23.8 & 0.388 \\
        Ours & \textbf{22.8} & \textbf{0.415} \\
        \hline
        \end{tabular}
    \end{minipage}\hfill
    \caption{Distribution-level evaluation on style transfer.
        (Left) Quantitative comparison on latent-guided synthesis.
        (Right) Quantitative comparison on reference-guided synthesis.}
    \label{tab:style}
\end{table}

\begin{figure}[t]
\centering
    \includegraphics[width=\linewidth]{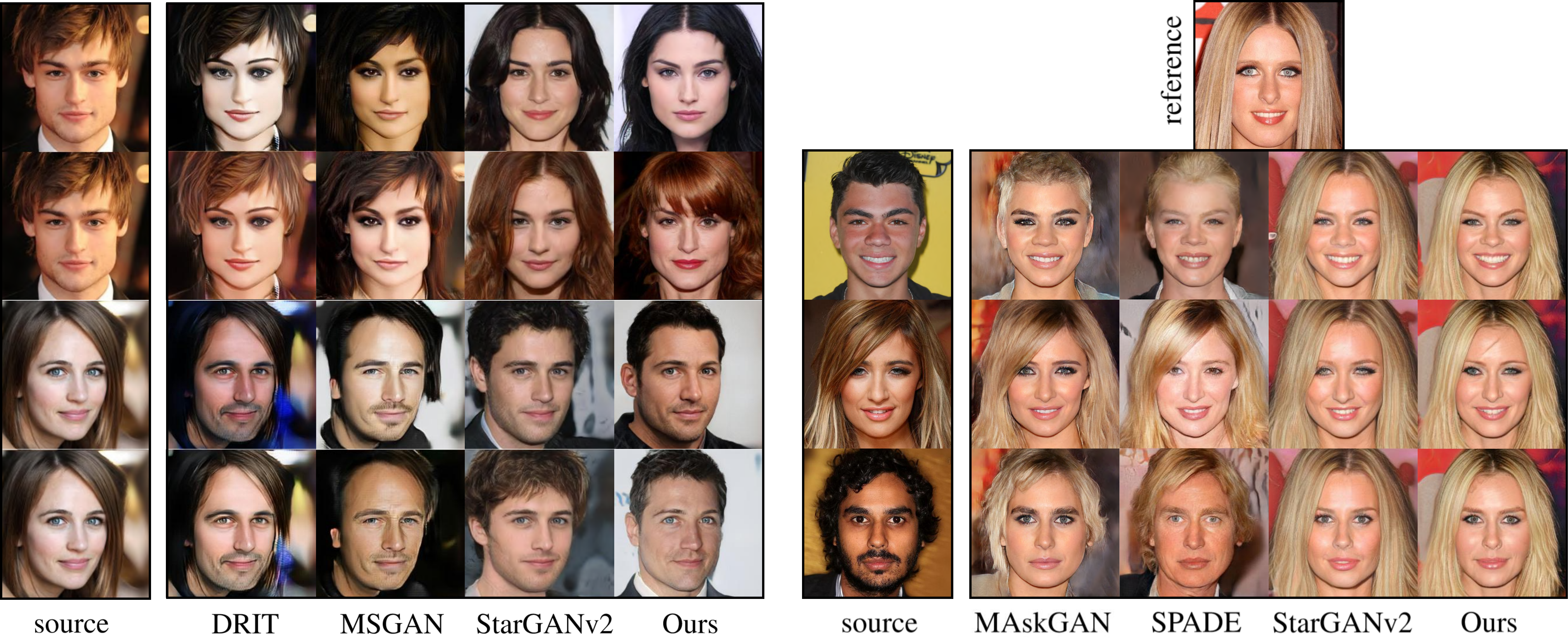}
   \caption{Qualitative comparison of style transfer on image synthesis.
   (Left) Latent-guided generation using random latent codes.
   (Right) Reference-guided generation.}
\label{fig:style}
\end{figure}

\subsection{Evaluation Metrics}
\noindent \textbf{Distribution-level Evaluation.} 
To evaluate diversity and visual quality, we use the Fréchet inception distance (FID) \cite{heusel2017gans} and learned perceptual image patch similarity (LPIPS) \cite{zhang2018unreasonable} metrics. 
\noindent \textbf{Attribute-level Evaluation.}
To evaluate the ability to manipulate target attributes, we train binary facial classifiers for the specific attributes on CelebA beforehand.
In particular, we use a ResNet-18 \cite{he2016deep} architecture.
\noindent \textbf{Segmentation-level Evaluation.}
To evaluate the capacity to generate synthetic images conditioned on the input mask, we train a facial semantic segmentation network on CelebA-HQ.
In particular, we use a U-Net \cite{ronneberger2015u} architecture that measures the pixel-wise accuracy between the input layout and the predicted parsing results.
\noindent \textbf{Identity-level Evaluation.}
In the context of style and attribute facial manipulation, it can be relevant to preserve the identity.
Therefore, we employ a pretrained face verification classifier on LFW \cite{huang2008labeled}.
In particular, we use ArcFace \cite{deng2019arcface} model with an accuracy of 99.5\%.

\subsection{Style transfer}
We start our experimental evaluation, assessing the style transfer ability of our model from two perspectives: latent-guided synthesis and reference-guided synthesis.

Latent-guided refers to the fact that the system learns to model random noise into valid latent code representations that account for a specific style.
\autoref{tab:style} provides a quantitative comparison of the baseline methods.
Our approach provides very competitive results, outperforming most of the models on both FID and LPIPS score, and being very close to \cite{choi2020stargan}.
The main reason for these results is the ability of to morphologically change the attributes, resulting in a wide variety of synthetic faces.
Our model produces highly diverse results given a single input, having a balanced image quality.
Furthermore, we conduct a visual inspection of a few samples.
In \autoref{fig:style}, a qualitative comparison between the different baselines is illustrated.
Each column contains the style transfer result from a different random  noise input.
The top two rows correspond to the results of converting male to female and vice versa in the bottom two rows.
We observe that both \cite{choi2020stargan} and our model generate images with a higher visual quality compared to \cite{lee2018diverse} and \cite{mao2019mode} model.
While most of the time \cite{lee2018diverse} synthesizes plausible outcomes, they do not contain morphological changes leading to poorer style transfers.
On the other hand, \cite{mao2019mode} generates results containing more substantial modifications, nonetheless, the method seems to fail to synthesize realistic images.

The second perspective, reference-guided, refers to the fact that the system learns to extract high-level semantics such as hairstyle, makeup, beard and age from the reference images, and to represent it in a latent code.
The pose and identity of the source images are preserved. 
\autoref{tab:style} shows the quantitative comparison of our method and the baseline methods for reference-guided synthesis.
Additionally, we also benchmark the models from \cite{park2019semantic} and \cite{lee2020maskgan}.
It should be noted, that these methods are using  reference labels to the sources.
In this second experiment, our model achieves superior scores in both FID and LPIPS metric compared to the competing models.
This implies that our approach produces the most diverse and realistic results while considering the styles of reference images.
\autoref{fig:style} compares the appearances of FacialGAN with the baseline methods.
We observe our approach and \cite{choi2020stargan} have successfully rendered distinctive styles, e.g., hairstyle, makeup and skin colour, while \cite{lee2020maskgan} fails at hairstyle translation, and \cite{park2019semantic} mostly matches only the colours of reference images. 

\begin{figure}[t]
\centering
    \includegraphics[width=\linewidth]{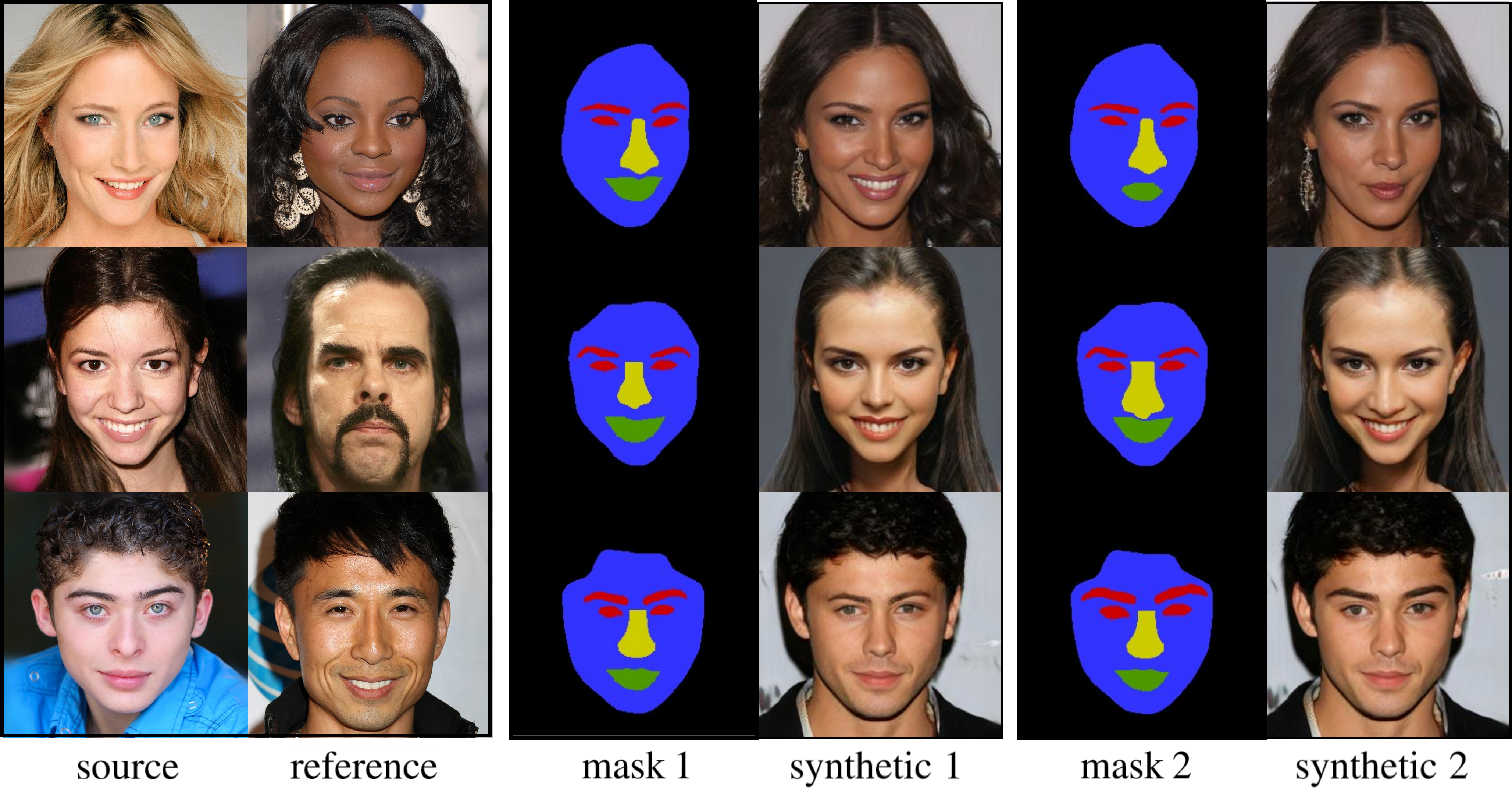}
   \caption{In this picture, we show that our model is able to learn to transform a source image to reflect the style of a given reference image while being consistent with the semantic mask. 
   The source and style reference images appear in the first two columns, whereas  the respective transformation masks are given in column 3 and 5. 
   The columns 4 and 6 show the generated images. Each row displays a modified attribute, i.e., mouth, nose and eyes (eyebrows).}
\label{fig:samples}
\end{figure}

\subsection{Attribute transfer}
On top of the style transfer capability, our model also allows a fine-grained attributes manipulation based on supervised signals.
Those signals come from one hot encoded vector that determine the gender, and from the semantic facial mask that controls the eyes, eyebrows, nose and mouth.
\autoref{fig:samples} shows synthesize images with style and attribute modifications.

We start our analysis by investigating the control over attribute transfers, in particular over the gender.
One important difference between some baseline models and ours, is how the gender information is encoded into the generative system.
On the one hand, we have \cite{park2019semantic} and \cite{lee2020maskgan} that employ the reference image to determine the gender.
In other words, they treat gender as a part of the style information.
On the other hand, we have our model and \cite{choi2020stargan} that use a label to set the gender, and therefore, treating it independently of the style.
The first column of \autoref{tab:attributes} shows the classification accuracy that each baseline achieves on a gender classifier when targeting only male outputs.
As one can expect, there is a clear difference between those approaches that use a gender-specific signal, and those who does not.
Note that the pretrained classifier has an accuracy of 96.1\% which will serve as a ground-truth.

Besides gender, our model allows manipulating eyes, eyebrows, nose and mouth independently, and synthesizing images accordingly.
As mentioned above, such control comes from the segmentation information of the masks.
Our approach is able to react to modifications on the segmentation mask at pixel-level.
Hence, it is possible to scale the size of a specific attribute, or even to redraw a new mask completely from scratch. 
The main limitation in terms of manipulation would arise from the need of realistic customized masks so that they could be translated into realistic faces.
We conduct an experiment where we choose \textit{smiling} attribute to compare with previous works \cite{park2019semantic,lee2020maskgan}.
Drawing smiles is a challenging task since not only affects the mouth attribute, but also influences the whole expression of the face, resulting in large geometry variety.
To run this evaluation, we manipulate non-smiling masks to be smiling and then, we generate a new set of images.
The second column of \autoref{tab:attributes} shows how our method achieve very competitive results, outperforming the baselines.

Moreover, we evaluate identity preservation. 
We first study the effect of style transfer on face recognition, where we measure the accuracy of face recognition of the source and the generated image with style transfer. 
We achieve an identity accuracy of 89.8\%.
Further, we apply attribute modification i.e., smiling faces, where we measure the accuracy of face recognition between the source and the generated image with both attribute modification and style transfer.
The accuracy in \autoref{tab:attributes} shows our method outperforms the baselines with an accuracy of 89.8\%.
Even with additional modification, our method is able to preserve identity better than the baselines.
Having a high attribute transfer accuracy is an important step towards our goal. 
Nevertheless, this metric might be incomplete as it does not evaluate the precision of the mask, or if the identity is preserved. 
Hence, we conduct a segmentation study per attribute, see \autoref{tab:segmentation}, where we assess the consistency between the input layout and the predicted parsing results in terms of pixel-wise accuracy.

\begin{table}[t]
    \begin{minipage}[b]{0.65\linewidth}
		\centering
        \begin{tabular}[b]{|l|c|c|c|}
            \hline
            Method & Male acc. & Smile acc. & Identity acc. \\
            \hline\hline
            SPADE \cite{park2019semantic} & 54.5 & 73.8 & 70.7\\
            MaskGAN \cite{lee2020maskgan} & 71.7 & 77.3 & 76.4\\
            StarGANv2 \cite{choi2020stargan} & \textbf{100} & - & -\\
            Ours & \textbf{100} & \textbf{81.4} & \textbf{89.8*}\\
            \hline
            Ground-truth & 96.1 & 92.3 & 99.5\\
            \hline
        \end{tabular}
        \caption{Second and third column show the attribute-level evaluation on male and smiling transfer synthesis accuracy. 
        Fourth, identity-level evaluation after drawing smiles.
        *Note that our synthetic image also contains style modifications, making more challenge the identity preservation.}
    \label{tab:attributes}
	\end{minipage}\hfill
    \begin{minipage}[b]{0.33\linewidth}
        \centering
        \begin{tabular}[b]{|l|c|c|}
        \hline
        Attribute &  Ours & GT \\
        \hline\hline
        Eyes & 98.39 &  98.81 \\
        Nose & 99.30 & 99.45 \\
        Mouth & 98.78 & 99.06 \\
        \hline
        All & 96.40 & 98.75 \\
        \hline
        \end{tabular}
        \caption{Segmentation-level evaluation to measure the consistency between the input mask and the predicted parsing results in terms of pixel-wise accuracy.}
        \label{tab:segmentation}
    \end{minipage}\hfill
\end{table}

\begin{figure}[t]
\centering
    \includegraphics[width=\linewidth]{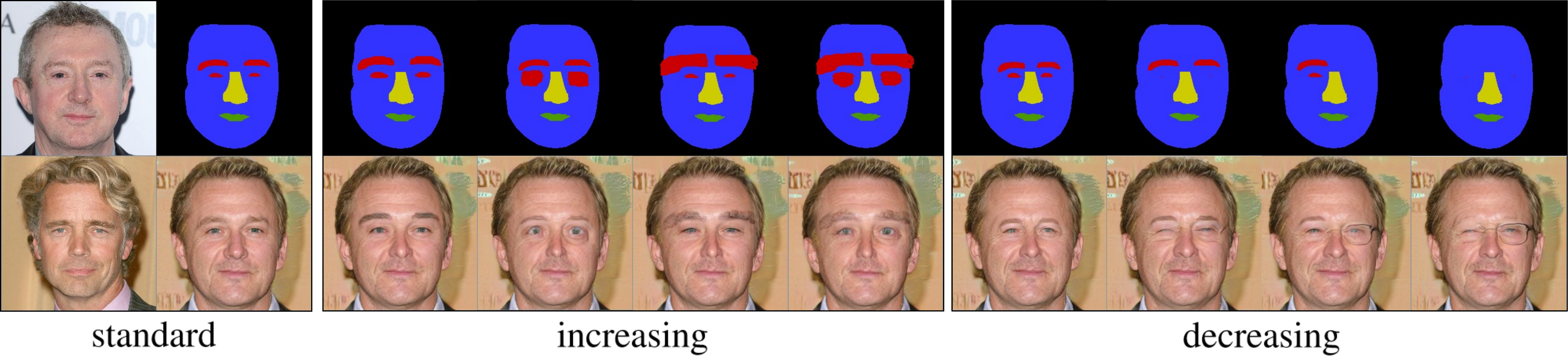}
   \caption{Results on extreme manipulation of eyes (eyebrows) mask.}
\label{fig:limitations}
\end{figure}

\subsection{Limitations}
We run an empirical study to determine under which circumstances our proposal starts to behave erroneously, producing inconsistent outputs. 
To find such limitations, we constantly increase the size of the mask of the target attributes and evaluate the results.
We repeat the same procedure but decreasing the mask size.
\autoref{fig:limitations} displays a few examples, where we can see how our model is following the mask if it can be translated into a realistic face.
Once the modified mask contains unnatural structures, e.g., no eyes, the network starts to ignore the mask input.
The main reason for this behaviour is the effect that the discriminator has over the generator during training, avoiding generating unrealistic faces.
A second limitation factor arises from our segmentation loss.
In order to generate informative gradients, it needs to work locally.
Therefore, we need training data with predefined areas where it will be applied, i.e., the mask of the target attributes.
Otherwise, the regions of no interest weaken the learning signal, leading to a loss of control of attribute editing.
\section{Conclusions}
We propose a novel interactive attribute manipulation and style transfers framework for facial image editing, coined FacialGAN.  
It learns to extract and to apply diverse style from a reference image while preserving the input face identity, and to incorporate geometry information of a guidance segmentation mask. 
A multi-objective strategy guarantees the generation of high-quality faces, and a fine-grained manipulation of facial attributes. 
Experimental results demonstrate that our FacialGAN outperforms state-of-the-art approaches. 
Finally, we provide an open-source end-user application for the evaluation of our model.

\bibliography{egbib}
\clearpage


\section{Supplementary Material}
\subsection{Direct Comparison with StarGANv2 and MaskGAN}
In this section, we present a detailed comparison between current state-of-the-art image-to-image facial translation models.
In particular, we benchmark our FacialGAN approach against StarGANv2 \cite{choi2020stargan} and  MaskGAN \cite{lee2020maskgan}.
\autoref{tab:comparison} shows a concise summary, emphasizing the main properties of each model.
\begin{table}[h]
	\centering
    \resizebox{\textwidth}{!}{\begin{tabular}[b]{|l|c|c|c|c|c|}
    \hline
    Method & scalability & style trans. & attr. manipulation & editing control & train stages \\
    \hline\hline
    StarGANv2 \cite{choi2020stargan} & limited & good & limited & none & one \\
    MaskGAN \cite{lee2020maskgan}  & good & limited & good & good & two \\
    Ours  & good & good & good & good & one \\
    \hline
    \end{tabular}}
    \caption{Summary of main differences between our model, StarGANv2 and  MaskGAN.}
    \label{tab:comparison}
\end{table}

We start outlying the differences of our proposal regarding StarGANv2 \cite{choi2020stargan}.
For the preservation of the target attributes, to guarantee the identity, \cite{choi2020stargan} employs a pretrained network based on an adaptive wing loss \cite{wang2019adaptive} that generates heatmaps.
These heatmaps, however, do not provide fine-grained control over the attributes.
They just signalize to the generator whether to keep or not all the facial attributes when applying the styles. 
Besides this limitation, the heatmaps require the input images to be vertically and horizontally aligned to have the eyes at the centre.
Otherwise, attribute preservation may fail.
Furthermore, the system relies on domain-specific modules to enforce semantic information, such as the output's gender.
As a result, the architecture and its performance are dependent on the number of different semantic labels with which the system works, limiting its scalability.
Overall, \cite{choi2020stargan} can successfully synthesize images of various styles, but with scalability issues and no pixel-wise control.
FacialGAN, on the other hand, can manipulate facial attributes at the pixel level while also applying styles from reference images.
To accomplish this, we have made a number of changes to our model.
First, we modify the generator to handle semantic mask labels as input.
This modification has two direct consequences: (1) it removes the heatmaps dependency, greatly simplifying the system, and (2) it allows the pixel information to be used to control the attributes.
To make use of such information, we add a segmentation network, a modified U-Net \cite{ronneberger2015u}, and a customized loss function that interacts with semantic input signals.
More specifically, we propose a new local segmentation loss that propagates informative gradients only from the region of interest, i.e. the target pixel-wise attributes.
In this way, we ensure that the generated output adheres to the mask specifications.
Finally, the addition of geometry information (segmentation mask) might be seen as an alternative to semantic information, allowing us to scale up our model to work with more attributes without adding complexity, i.e. dedicated modules.

The differences between FacialGAN and MaskGAN \cite{lee2020maskgan} are also notable.
Both models are radically different in terms of design.
\cite{lee2020maskgan} trains in a fairly complex setup, which is divided into two stages with three different networks.
In the first stage, the model learns the mapping between the semantic mask and the output image.
Once it has converged, they start training the second stage, where the method learns to model the user editing behaviour when manipulating semantic masks.
Additionally, there is an encoder-decoder architecture called MaskVAE, which  is in charge of generating geometrical masks for training, that needs to be pretrained beforehand.
All the training relies on the optimization of an adversarial, feature and perceptual loss.
Despite synthesizing successfully images with different pixel-wise control, \cite{lee2020maskgan} is not designed for advanced style transfers.
In other words, it lacks the ability to apply morphological changes when applying style, resulting in a very limited approach to style transfer.
FacialGAN, on the other hand, is capable of extracting and applying cutting-edge style transfers, as well as modifying the geometry of the image if needed.
We accomplish this by using a one-step training method in which the model learns the style while manipulating the semantic masks.
As a result, FacialGAN has a more compact training that produces superior distribution-level metrics (FID and LPIPS).
Identity preservation and attribute manipulation are two other areas where our proposal outperforms \cite{lee2020maskgan}.
Mainly due to our new local segmentation loss, that forces the generator to focus only on the target regions, leaving the rest unmodified.
Thus, the outcomes preserve unaltered the main features, respecting the identity, with pixel-wise control in case the user desires to change some attributes.

\subsection{Facial Editing Toolbox}
We propose an interactive facial toolbox that allows easy manipulation of both styles and attributes. 
The user chooses the source and reference images, and our model generates the desired combination in the desired direction.
In addition, the user can change the default mask, and the changes are reflected in the output.
We believe that such a tool can be very useful for validating results, allowing practitioners to continue to improve.
The source codes, pretrained models, and facial editing toolbox can be found on  \href{https://github.com/cc-hpc-itwm/FacialGAN}{Github}.
We also provide a video tutorial where we show how to use the toolbox.
It is available on \href{https://www.youtube.com/watch?v=N4jRSNKPB0s}{Youtube}.

\subsection{Additional Results}
\begin{figure}[h]
\centering
    \includegraphics[width=\linewidth]{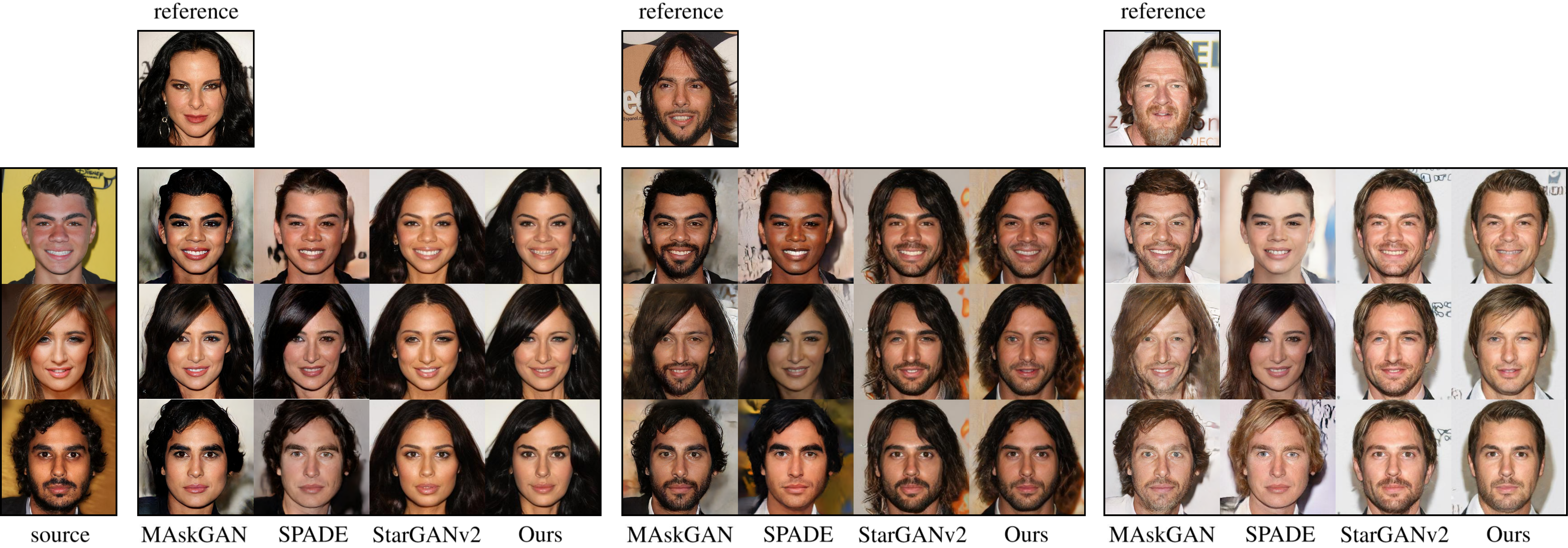}
   \caption{Qualitative comparison of style transfer on reference-guided generation.}
\label{fig:style2}
\end{figure}

\begin{figure}
\centering
    \includegraphics[width=\linewidth]{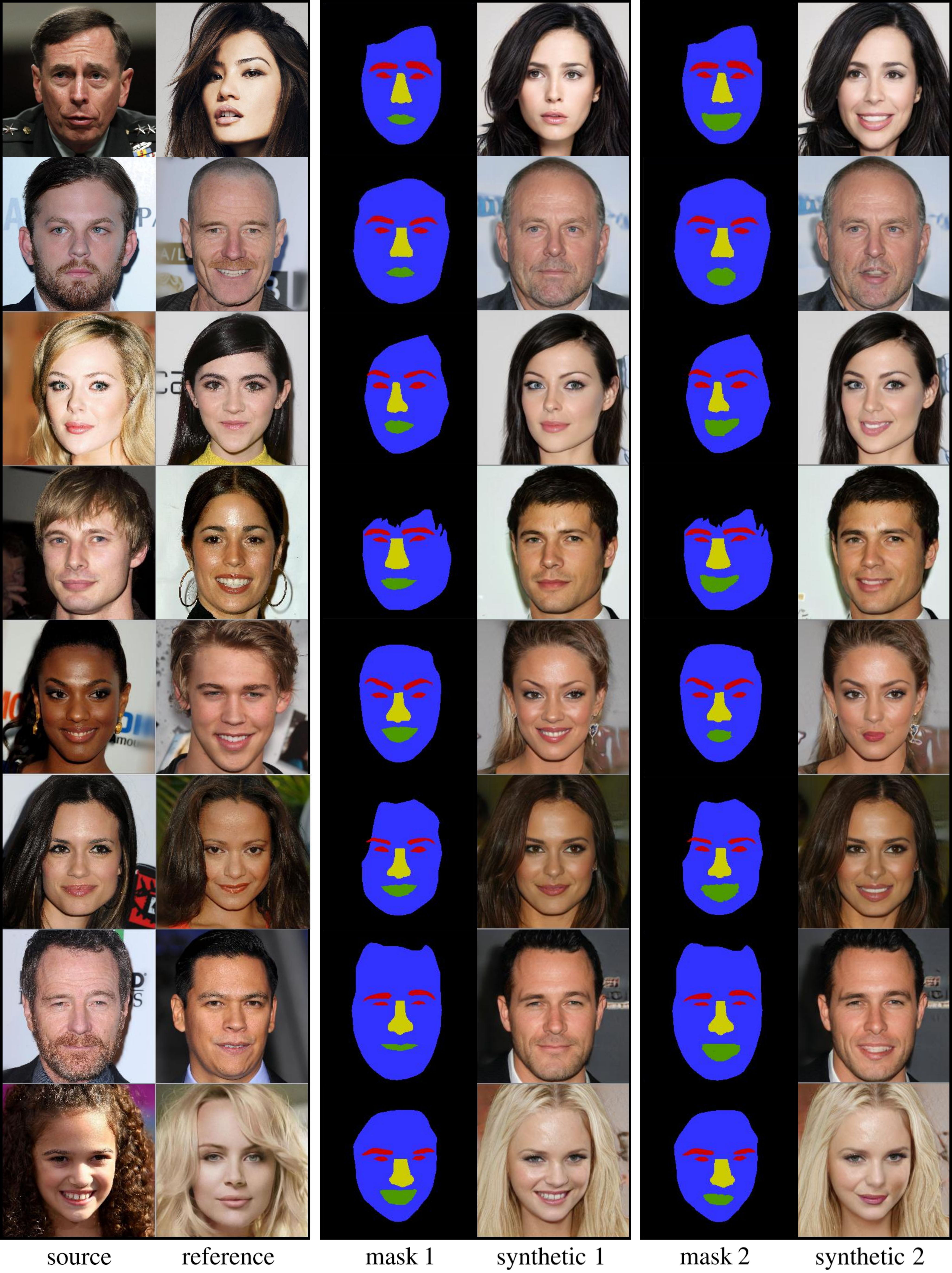}
   \caption{In this picture, we show that our model is able to learn to transform a source image to reflect the style of a given reference image while being consistent with the semantic mask. 
   In particular, we only modify the mask of the \textit{mouth}.
   The source and style reference images appear in the first two columns, whereas  the respective transformation masks are given in column 3 and 5.
   The columns 4 and 6 show the generated images.}
\label{fig:mouth}
\end{figure}

\begin{figure}
\centering
    \includegraphics[width=\linewidth]{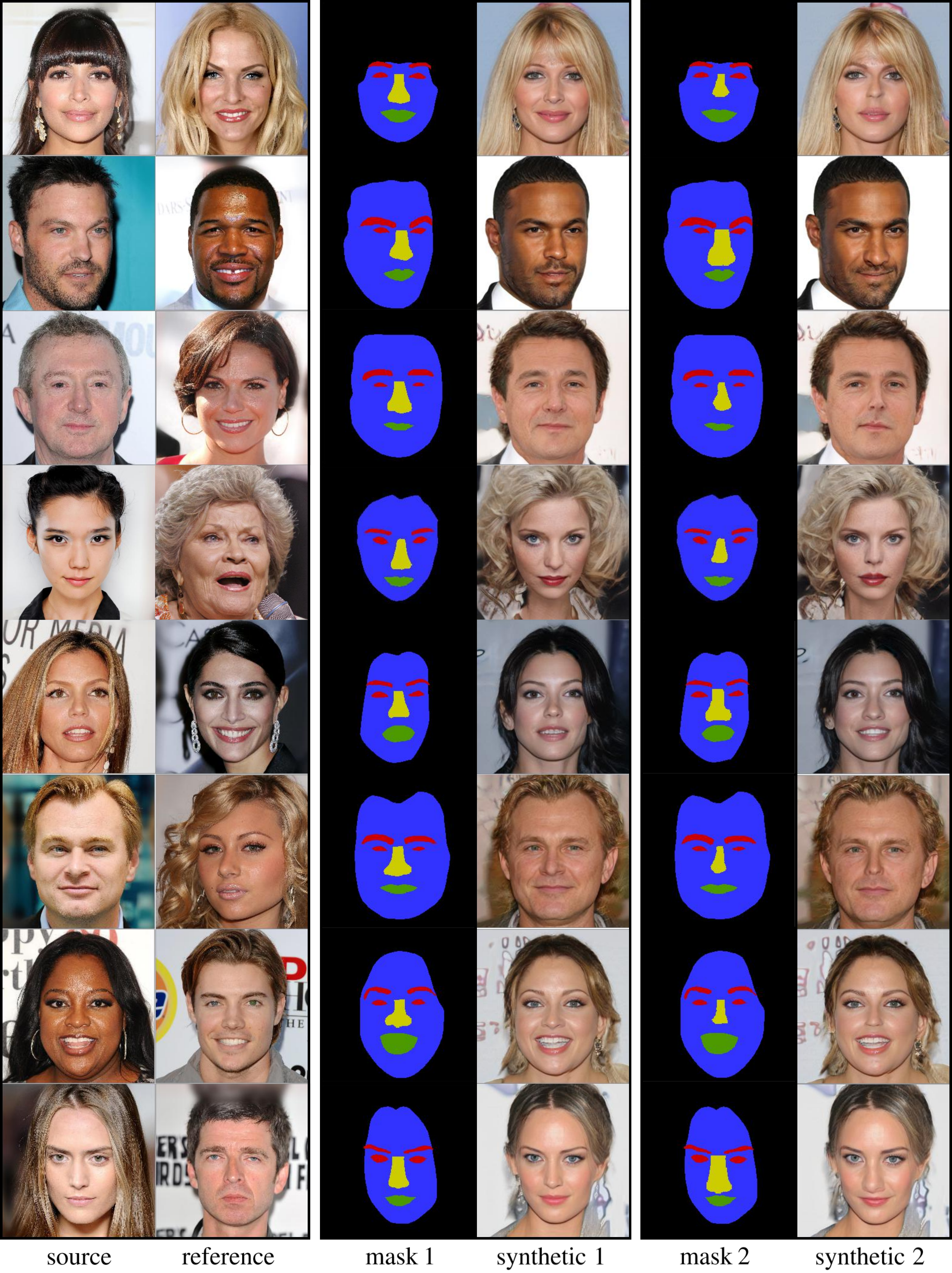}
   \caption{Our model is able to learn to transform a source image to reflect the style of a given reference image while being consistent with the semantic mask. 
   In particular, we only modify the mask of the \textit{nose}.
   The source and style reference images appear in the first two columns, whereas  the respective transformation masks are given in column 3 and 5. 
   The columns 4 and 6 show the generated images.}
\label{fig:nose}
\end{figure}

\begin{figure}
\centering
    \includegraphics[width=\linewidth]{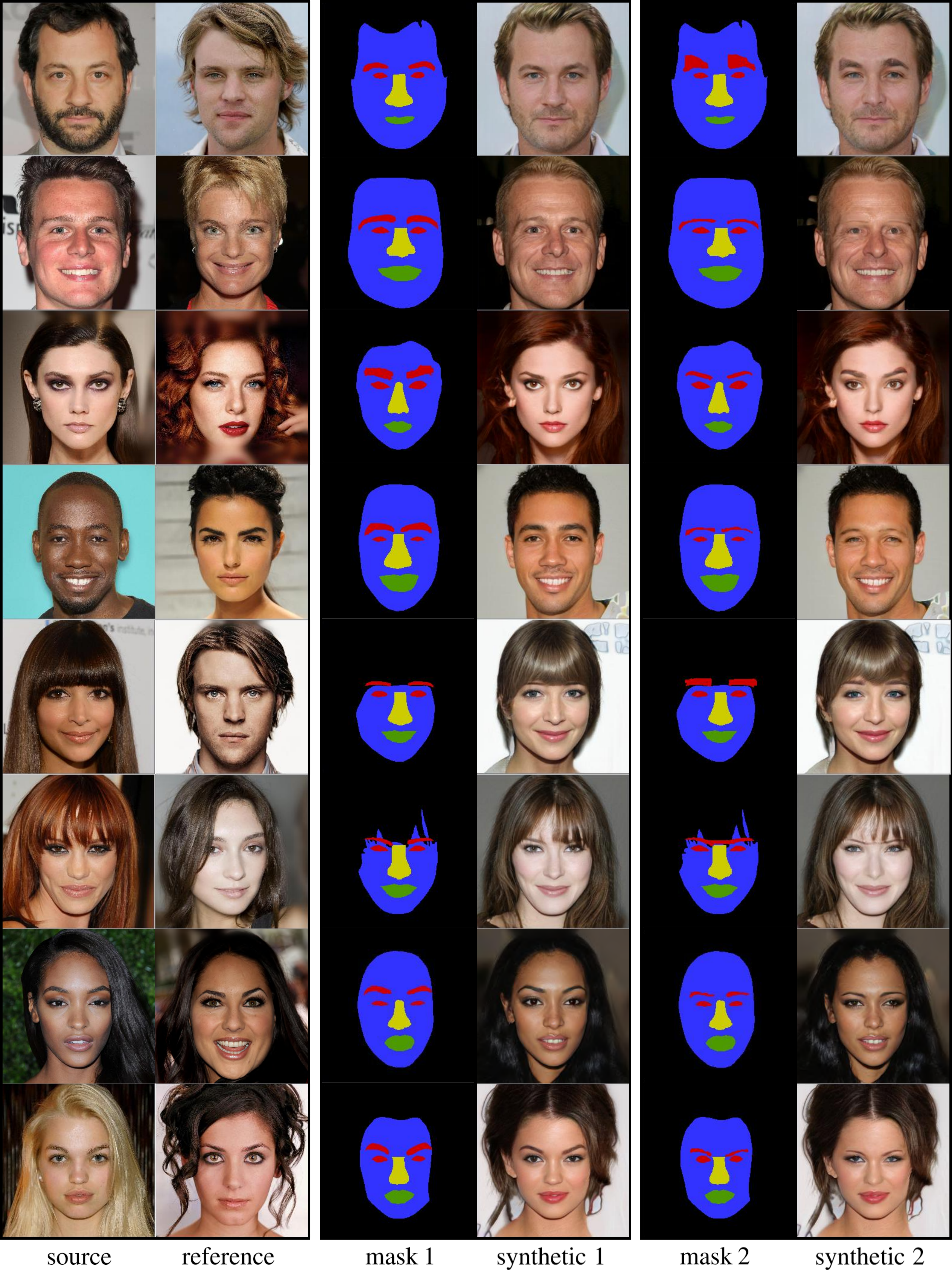}
   \caption{Our model is able to learn to transform a source image to reflect the style of a given reference image while being consistent with the semantic mask. 
   In particular, we only modify the mask of the \textit{eyes} (\textit{eyebrows}).
   The source and style reference images appear in the first two columns, whereas  the respective transformation masks are given in column 3 and 5. 
   The columns 4 and 6 show the generated images.}
\label{fig:eyes}
\end{figure}

\begin{figure}
\centering
    \includegraphics[width=\linewidth]{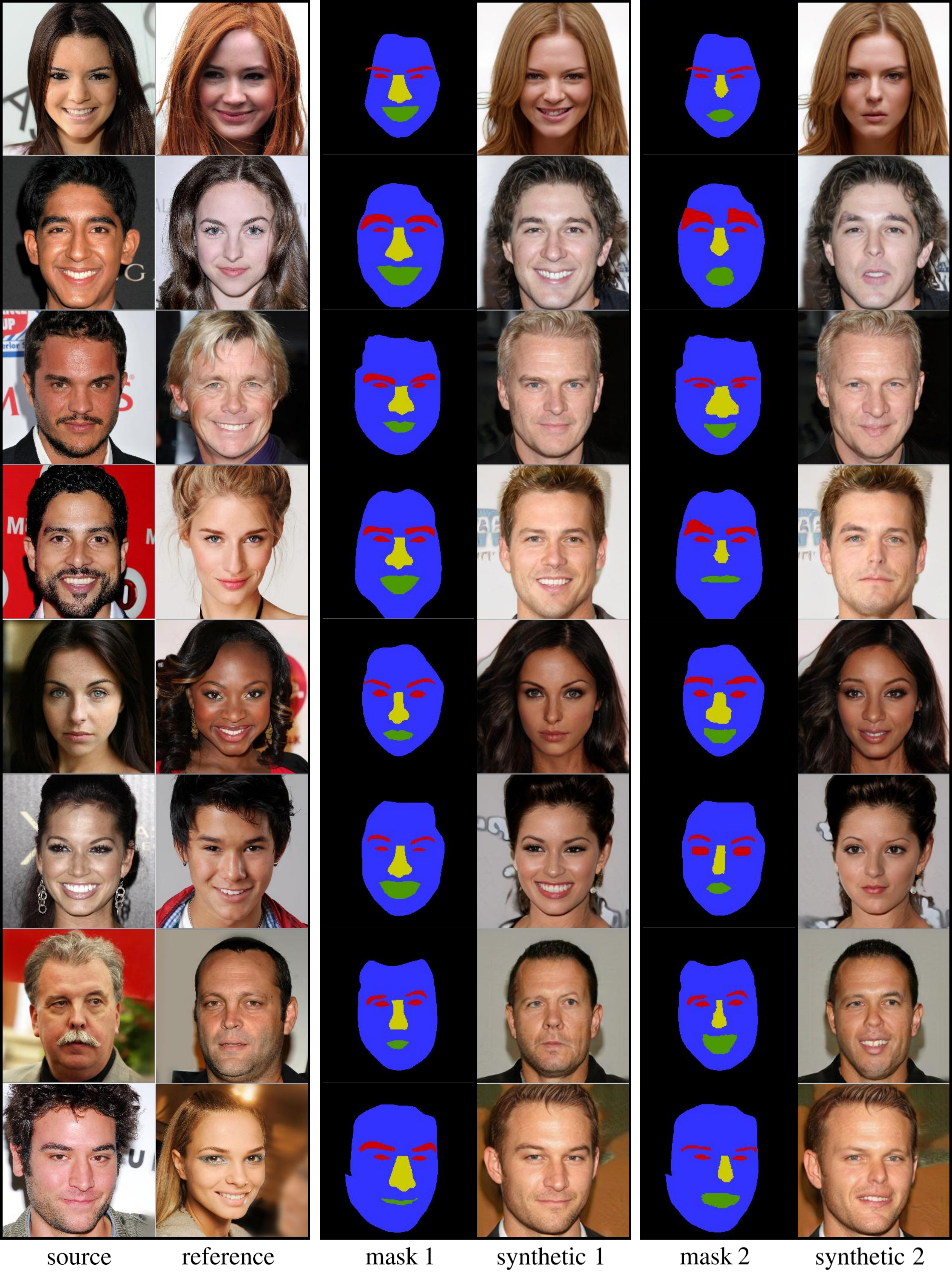}
   \caption{Our model is able to learn to transform a source image to reflect the style of a given reference image while being consistent with the semantic mask. 
   In particular, we modify more than one attribute of the mask.
   The source and style reference images appear in the first two columns, whereas  the respective transformation masks are given in column 3 and 5. 
   The columns 4 and 6 show the generated images.}
\label{fig:mix}
\end{figure}

\begin{figure}
\centering
    \includegraphics[width=\linewidth]{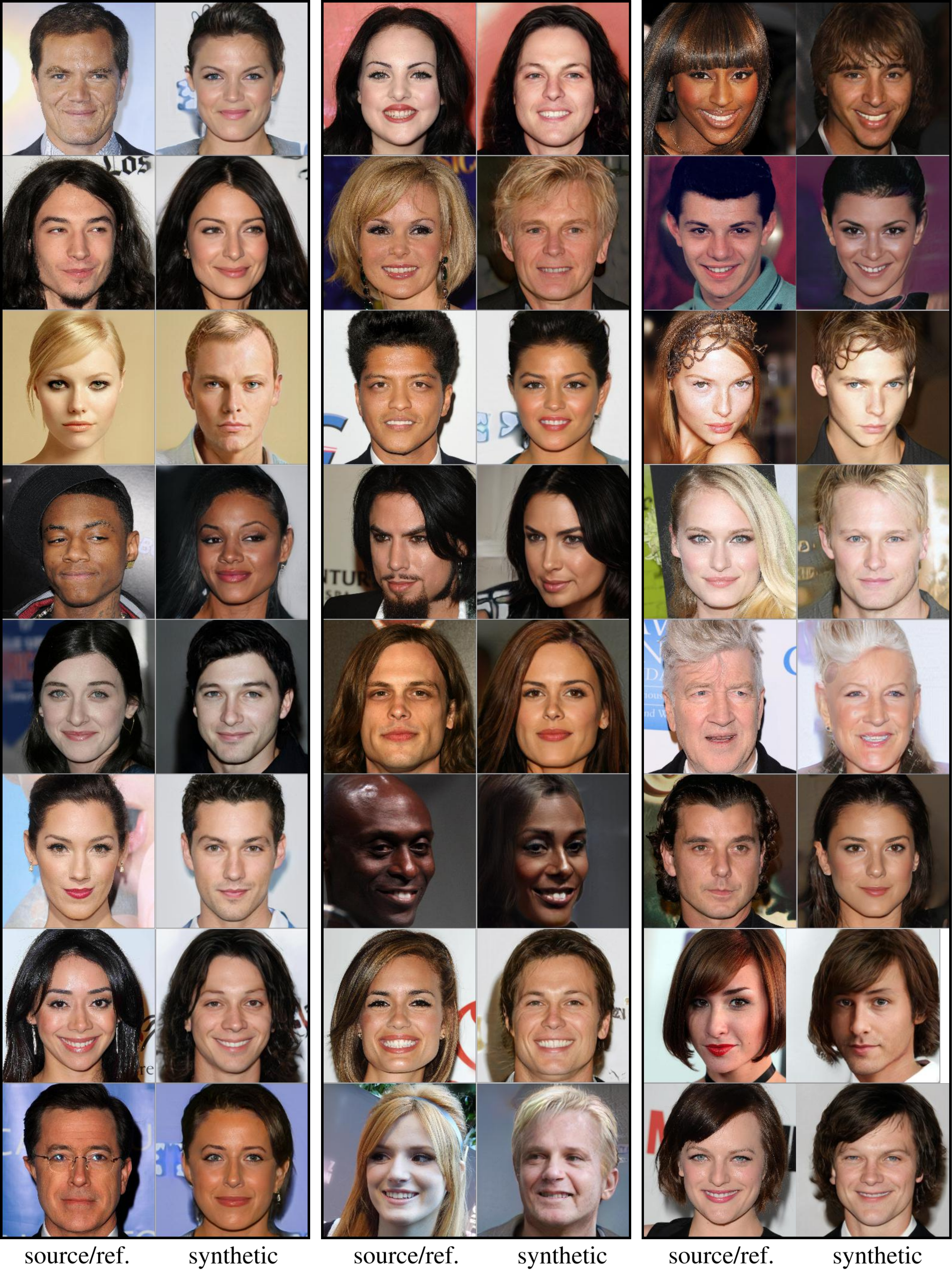}
   \caption{Our model is able to apply gender translation.
   The source and style reference images are the same, and appear in the first columns of each block. 
   The respective synthetic results from the gender transformation are given in second columns.}
\label{fig:gender}
\end{figure}

\autoref{fig:style2} displays additional comparisons with the baseline models on reference-guided generation.
Furthermore, we provide additional image synthesis results where we apply an extensive variety of styles and mask's modifications.
\autoref{fig:mouth}, \autoref{fig:nose} and \autoref{fig:eyes} display generated results when mouth, nose or eyes (eyebrows) have been altered through their segmentation masks. 
The results, where more than one attribute was changed, are shown in \autoref{fig:mix}.
Finally, we run a gender translation experiment where the source and the reference images are the same.
This way, if we exchange the original gender, we still get the ``same'' person with the same style, but with the opposite gender.
Results are displayed in \autoref{fig:gender}.

\end{document}